\documentclass[letter, 10pt, conference]{ieeeconf}     
\IEEEoverridecommandlockouts                              

\usepackage{cite}
\usepackage{amsmath,amssymb,amsfonts}
\usepackage{algorithmic}
\usepackage{graphicx}
\usepackage{textcomp}
\usepackage{xcolor}
\usepackage{caption}
\usepackage[switch,pagewise,columnwise]{lineno}
\usepackage{soul}
\usepackage{booktabs}
\usepackage[maxfloats=256]{morefloats}
\usepackage[ruled, vlined]{algorithm2e}
\usepackage{makecell}
\usepackage{subcaption}
\usepackage{gensymb}

\title{\LARGE \bf
Distributional Soft Actor-Critic with Harmonic Gradient for Safe and Efficient Autonomous Driving in Multi-lane Scenarios
}

\author{Feihong Zhang$^{1\#}$, Guojian Zhan$^{1\#}$, Bin Shuai$^{1}$, Tianyi Zhang$^{1}$, Jingliang Duan$^{2}$, 
Shengbo Eben Li$^{1, 3*}$
\thanks{This work is supported by NSF China with U20A20334 and Tsinghua University-Toyota Joint Research Center for AI Technology of Automated Vehicle. The first two authors contributed equally. All correspondences should be sent to S. E. Li with email: {\tt\small lishbo@tsinghua.edu.cn}.}
\thanks{
$^{1}$School of Vehicle and Mobility, Tsinghua
University, Beijing, 100084, China. 
$^{2}$School of Mechanical Engineering, University of Science and Technology Beijing, Beijing, 100083.
$^{3}$College of Artificial Intelligence, Tsinghua University, Beijing, 100084, China.}
}

\begin{document}

\maketitle
\thispagestyle{empty}
\pagestyle{empty}

\begin{abstract}
Reinforcement learning (RL), known for its self-evolution capability, offers a promising approach to training high-level autonomous driving systems. However, handling constraints remains a significant challenge for existing RL algorithms, particularly in real-world applications.
In this paper, we propose a new safety-oriented training technique called harmonic policy iteration (HPI). At each RL iteration, it first calculates two policy gradients associated with efficient driving and safety constraints, respectively. Then, a harmonic gradient is derived for policy updating, minimizing conflicts between the two gradients and consequently enabling a more balanced and stable training process. Furthermore, we adopt the state-of-the-art DSAC algorithm as the backbone and integrate it with our HPI to develop a new safe RL algorithm, DSAC-H. Extensive simulations in multi-lane scenarios demonstrate that DSAC-H achieves efficient driving performance with near-zero safety constraint violations.



\end{abstract}

\section{INTRODUCTION}
Autonomous driving (AD) has emerged as a transformative technology with the potential to revolutionize the transportation system, enhancing energy efficiency, and improving mobility accessibility for individuals \cite{li2022decision}.
However, the real-world deployment of autonomous vehicles is confronted with significant challenges, particularly in ensuring safety across diverse and unpredictable driving scenarios \cite{jiang2023reinforcement}. 

Traditional methods, including rule-based systems and imitation learning, have demonstrated limitations in scalability and adaptability, often requiring considerable high-quality data and failing to generalize to unseen scenarios\cite{duan2020hierarchical}. 
Reinforcement learning (RL) has gained significant attention as a powerful framework for addressing complex decision-making problems in autonomous driving \cite{li2023rlbook}. Unlike traditional methods, RL enables agents to learn optimal policies through interaction with the environment, aiming to maximize expected cumulative rewards \cite{zhan2024transformation, shuai2024optimal}. 
As a pioneering work, Wayve corporation demonstrated the first application onto a fully sized autonomous vehicle\cite{wayve}. With a monocular camera image and steering angle as the observation, they successfully trained a driving policy on the on-board computers to generate the expected steering and speed, which could conduct the lane following on a 250 meter section of road. Maramotti \emph{et al.} trained a neural network using the data processed by localization and perception algorithms on board of the self-driving car, which can predict both acceleration and steering angle to drive in obstacle-free urban environments \cite{maramotti2022tackling}.
Duan \emph{et al.} introduced the digital twin simulation system to realize a more complex experiment, in which the ego vehicle rode on a real intersection, and the surrounding vehicles were generated by a virtual perception system\cite{duan2021encoding}. {This methodology was later adopted by many researchers to enhance the validity verification of RL algorithm in real-world scenarios\cite{ma2021model,voogd2023reinforcement}.}
Recently, Guan \emph{et al.} proposed the integrated decision and control (IDC) to handle driving on intersection roads, wherein they calculated the policy gradient using a prior  dynamic model for ego vehicle and a navie prediction model for surrounding vehicles \cite{guan2022integrated}. Although they achieve relatively realistic driving effects, their reliance on prior models inevitably leads to significant conservatism. Overall, it is non-trivial to develop a RL-driven autonomous driving system, primarily due to the challenging safety constraints with other road users.

The urgent demand of safety has driven the rapid development of safe reinforcement learning (Safe RL). To tackle constraints, researchers have integrated various constrained optimization techniques into standard RL algorithms, which are also known as constrained RL algorithms. In the early stages, penalty methods, such as interior-point and exterior-point, were widely adopted due to their ease of implementation \cite{ren2022self, yoo2021dynamic}. To adaptively balance the reward and safety, recent research has majorly turned to the Lagrange multiplier method \cite{paternain2019learning}.
For each constraint term, this approach leverages a scalar called Lagrange multiplier, to integrate it into the primal reward objective and finally obtain a scalar criterion called Lagrangian. By further employing dual-descent-ascent training technique, this method explores the solution space of both policy parameters and multiplier to strike a balance between maximizing reward performance and satisfying safety constraints \cite{pmlr-v119-stooke20a}. Nevertheless, properly managing the multiplier update remains a challenge in this minimax training paradigm and is prone to causing significant oscillations. 
In conclusion, regardless of the specific
constraint optimization techniques used to enable RL for handling constraints, the policy inevitably will face the challenge of balancing reward and safety objectives \cite{xu2021crpo}.

In this paper, we introduce a safety-oriented training technique, harmonic policy iteration (HPI), designed to tackle the potential  conflicts between reward and safety objectives. Specifically, during each iteration of the RL process, two policy gradients associated with efficient driving and safety constraints are computed. Subsequently, a harmonic gradient is derived by minimizing conflicts with both of these gradients, thereby facilitating a more balanced and stable training procedure. We integrate this technique with the state-of-the-art DSAC algorithm \cite{DSAC, DSACT}, giving rise to a new safe RL algorithm named DSAC-H. Through extensive simulations conducted in multi-lane scenarios, DSAC-H is proven to be capable of achieving efficient driving performance while maintaining near-zero safety constraint violations.

\section{Preliminary}

In this section, we first introduce the principles of RL and Safe RL. Next, we present the DSAC algorithm, which stands out as one of the state-of-the-art model-free RL algorithms and serves as the backbone of our proposed DSAC-H.

\subsection{Principles of RL and Safe RL}

RL is a type of machine learning where an agent learns to make decisions by interacting with an environment, with the goal of maximizing the cumulative reward over time. The standard RL setting can be formalized as a Markov decision process (MDP), defined by a tuple \((\mathcal{S}, \mathcal{A}, P, r, \gamma)\), where \(\mathcal{S}\) is the state space, \(\mathcal{A}\) is the action space, \(P: \mathcal{S} \times \mathcal{A} \times \mathcal{S} \to [0, 1]\) is the transition probability function, \(r: \mathcal{S} \times \mathcal{A} \to \mathbb{R}\) is the reward function, and \(\gamma \in [0, 1)\) is the discount factor. The agent's policy \(\pi: \mathcal{S} \to \mathcal{A}\) maps states to actions, with the aim of maximizing the expected cumulative reward:
\begin{equation}
J_{r} = \mathbb{E}_{\tau \sim \pi} \left[ \sum_{t=0}^{\infty} \gamma^t r(s_t, a_t) \right].
\end{equation}
where \(\tau = (s_0, a_0, s_1, a_1, \ldots)\) is a trajectory generated by following policy \(\pi\).

Safe RL extends standard RL by incorporating safety constraint, which is expressed as
\begin{equation}
J_c = \mathbb{E}_{\tau \sim \pi} \left[ \sum_{t=0}^{\infty} \gamma^t c(s_t, a_t) \right] \leq 0
\end{equation}
where \(c: \mathcal{S} \times \mathcal{A} \to \mathbb{R}\) is the constraint function. The goal of Safe RL is to find a policy \(\pi\) that maximizes the expected cumulative reward while ensuring that the expected cumulative cost does not exceed zero. When the cost signal is defined as non-negative, it means the constraint must be strictly satisfied at each step.

\subsection{Distributional Soft Actor-Critic}

DSAC combines the strengths of maximum entropy RL and distributional RL framework, and achieves the state-of-the-art performance on the publicly available MuJoCo benchmark \cite{DSAC, DSACT}.
Technically, 
DSAC  treats the return, i.e., cumulative discounted rewards, as a random variable $Z$ and directly learns its distribution rather than just its expected value as
\begin{equation}
\begin{aligned}
Z^{\pi}(s_t,a_t) &= r_t + \sum^{\infty}_{i=t+1} \gamma^{i-t} \left[ r_i - \alpha \log \pi(a_i|s_i) \right], \\
    Q^{\pi}(s_t,a_t) &= \mathbb{E}[Z^{\pi}(s_t,a_t)].
\end{aligned}
\end{equation}
This technique allows DSAC to reduce the longstanding value overestimation issue compared to traditional methods that focus solely on the expected value, leading to a more accurate learned value function.

DSAC also incorporates the principles of maximum entropy RL to maximize both the expected accumulated reward and the entropy of the policy. The objective function is given by:
\begin{equation}
\label{eq.policy_objective}
J_{r} = \mathop{\mathbb{E}}_{(s_i,a_i)\sim\rho_{\pi}}\Big[\sum^{\infty}_{i=t}\gamma^{i-t} \left[ r_i + \alpha\mathcal{H}(\pi(\cdot|s_i)) \right] \Big],
\end{equation}
where $\mathcal{H}(\pi(\cdot|s_i))$ is the policy entropy, \(\alpha\) is the temperature parameter that balances the importance of the entropy term relative to the reward, controlling the stochasticity of the optimal policy. 
The maximum entropy objective encourages more exploration than the standard RL objective.

To conclude, the policy evaluation (PEV) step of DSAC is achieved by the distributional form Bellman operator:
\begin{equation}
\nonumber
\mathcal{T^{\pi}}Z^\pi(s,a) \overset{D}{=} r(s,a) + \gamma \left( Z^\pi(s',a') - \log \pi(a'|s') \right),
\end{equation}
where \(\overset{D}{=}\) indicates that two random variables have the same probability distribution. The return distribution is practically optimized by minimizing the distributional distance between the Bellman-updated and current return distributions, which is expressed as:
\begin{equation}
\label{eq.policy_eva}
Z_{\rm{new}} =  \arg\min_{Z} \mathop{\mathbb{E}}_{(s,a)\sim\rho_{\pi}}\left[ d\left(\mathcal{T}^{\pi}_{\mathcal{D}}Z_{\rm{old}}(\cdot|s,a), Z(\cdot|s,a)\right) \right],
\end{equation}
where \(d\) is the Kullback-Leibler (KL) divergence.

In the policy improvement (PIM) step, a new policy is found that is better than the current policy \(\pi\) as
\begin{equation}
\label{eq.policy_imp}
\begin{aligned}
\pi_{\rm{new}} = \arg\max_{\pi} \mathbb{E}_{(s,a)\sim\rho_{\pi}}\left[ Q^{\pi}(s,a) - \alpha \log \pi(a|s) \right].
\end{aligned}
\end{equation}
By alternatively performing the PEV and PIM steps in \eqref{eq.policy_eva} and \eqref{eq.policy_imp}, we can gradually approach the optimal policy and value functions.

\section{Method}

In this section, we first introduce the safety-oriented training technique, harmonic policy iteration (HPI), and then present the DSAC-H algorithm, which leverages HPI to enhance the search for a feasible and optimal policy.

\subsection{Harmonic Policy Iteration}

We extend the DSAC algorithm to handle safety constraints by learning two separate value functions: \(Q_r(s,a)\) for the reward and \(Q_c(s,a)\) for the safety constraint. Each value function updating follows a standard policy evaluation manner, enabling the agent to evaluate the expected cumulative reward and cumulative cost independently. The return distributions for rewards and constraints are defined as:
\begin{equation}\label{eq:reward_return}
Z_r^\pi(s_t,a_t) = r_t + \sum_{i=t+1}^{\infty} \gamma^{i-t} \left[ r_i - \alpha \log \pi(a_i|s_i) \right],
\end{equation}
\begin{equation}\label{eq:constraint_return}
Z_c^\pi(s_t,a_t) = c_t + \sum_{i=t+1}^{\infty} \gamma^{i-t} c_i.
\end{equation}
Their corresponding value functions are:
\begin{equation}\label{eq:value_functions}
Q_r^\pi(s,a) = \mathbb{E}[Z_r^\pi(s,a)], \quad Q_c^\pi(s,a) = \mathbb{E}[Z_c^\pi(s,a)].
\end{equation}
Thanks to the distributional form value function, HPI enables precise evaluation of both values during training. This allows the agent to accurately assess the expected cumulative reward and cost, which is crucial for identifying a feasible and optimal policy.

As for policy improvement, existing Safe RL algorithms typically combine reward and safety objectives using a single scalar parameter, such as a penalty factor or Lagrange multiplier, to form a weighted sum. However, this factor generally needs to be carefully tuned and these two objectives are highly likely to conflict with each other.

The policy parameters are denoted by $\theta$, the gradients of the reward and safety objectives are expressed as:
\begin{equation}\label{eq:gradients}
\begin{aligned}
    g_r &= \nabla_\theta J_r = - \nabla_\theta [Q_r - \alpha \log \pi], \\
    g_c &= \nabla_\theta J_c = \nabla_\theta Q_c.
\end{aligned}
\end{equation}
When these gradients conflict (i.e., \(\langle g_r, g_c \rangle < 0\)), directly combining them using a weighted sum may degrade performance in one or both objectives. In cases where conflicts arise, the inner product of \( g_r \) and \( g_c \) is negative, indicating that the angle between them is obtuse. Geometrically, this means that any movement in the direction of \( g_r \) degrades \( J_c \), while any movement in the direction of \( g_c \) degrades \( J_r \), regardless of the step size. This mutual interference makes it challenging to improve both objectives simultaneously. On the other hand, when no conflict exists, the inner product of \( g_r \) and \( g_c \) is positive, meaning they form an acute angle. In this scenario, a properly chosen step size allows movement along either gradient without adversely affecting the other objective, enabling smoother optimization.

\begin{figure}[htbp]
  \centering
  \includegraphics[width=1.25\linewidth, trim={7.2cm 0.0cm 0.0cm 0.0cm}, clip]{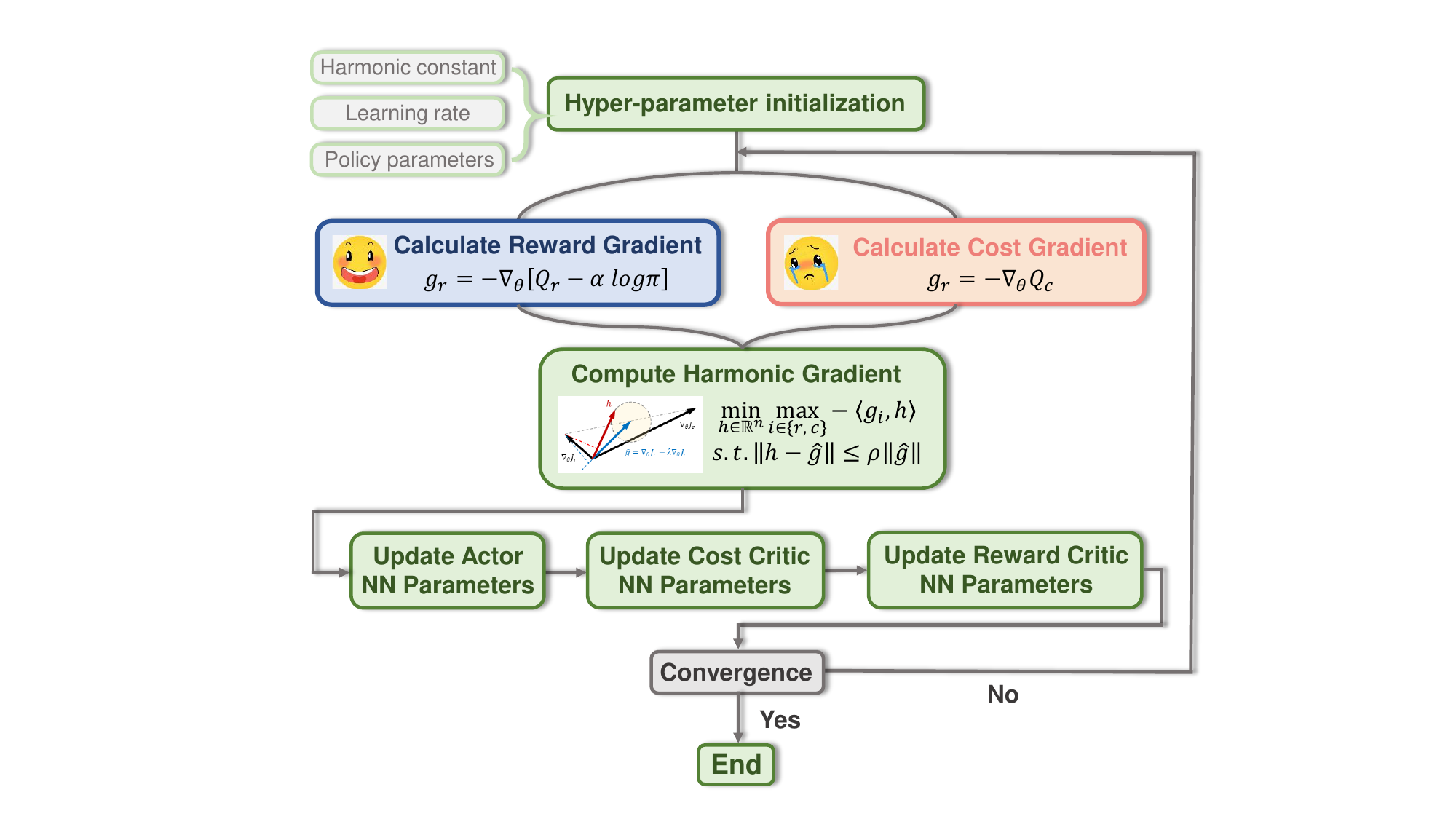}
  \caption{Harmonic Policy Iteration Framework.}
  \label{fig.Framework}
\end{figure}

HPI framework is presented in Figure \ref{fig.Framework}. Our HPI addresses this issue by combining the reward and safety objectives at the gradient level in an optimal manner. 
To mitigate gradient conflicts and promote simultaneous improvement in both reward and safety performance, we formulate the inner product value as the optimization objective. Specifically, we design a minimax optimization problem at each iteration to compute a harmonic gradient \( h \in \mathbb{R}^n \) for policy updating. Since gradient estimation in RL is typically performed using small-batch sampling, and the gradient magnitude must be controlled to ensure stable training, we introduce a trust region constraint. This constraint is defined as a closed ball, centered at a nominal gradient \( \hat{g} = g_r + \lambda g_c \), where \( g_r \) and \( g_c \) represent the gradients for reward optimization and safety constraints, respectively. The radius of this ball is adaptively determined based on the magnitude of \( \hat{g} \), ensuring that the optimization remains within a reasonable update range. The complete formulation is given by

\begin{equation}\label{eq:harmonic_gradient}
\begin{aligned}
\min_{h \in \mathbb{R}^n} \max_{i \in \{r, c\}} &  -\langle g_i, h \rangle \\
\text{s.t.}  \quad \| h - \hat{g} \| & \leq \rho \| \hat{g} \|, \\
\end{aligned}
\end{equation}
where \(\rho \in [0,1)\) is the harmonic constant controlling the trust region size. The minimax problem \eqref{eq:harmonic_gradient} consists of two iterative loops: an inner loop that identifies the worst-case objective and an outer loop that optimizes its inner product within the trust region.  

The solution \(h\) ensures that the policy update improves both reward and safety simultaneously with minimal gradient conflict by maximizing the inner products of their individual gradients with \(h\). When \( \rho > 0 \), it determines the extent to which \( h \) can be adjusted to mitigate gradient conflicts. A value of \( \rho = 0 \) implies that \( h \) is strictly equal to the nominal gradient \( \hat{g} \), allowing no adjustment.

\subsection{DSAC-H Algorithm}
During training process, the harmonic gradient \(h\) is computed at each iteration and used for policy updating, enabling HPI to balance reward maximization and constraint satisfaction. This approach ensures that the policy update direction is optimized for both objectives, leading to improved safety and performance. 
The detailed pseudocode of DSAC-H is presented in Algorithm \ref{alg:dsac-h}.

\begin{algorithm}[ht]
   \caption{DSAC-H}
   \label{alg:dsac-h}
   
    \KwIn{Harmonic constant $\rho_0 \in [0, 1)$, learning rate $\eta_1 > 0$ for actor, learning rate $\eta_2 > 0$ for two critics,  parameters $\theta_0$, $\omega_0$, $\phi_0$ for policy, value, and cost value } 
        \For{each training step $k$}
        {
              {\bfseries Update actor} \\
              Compute policy gradients $g_r, g_c$ \\
              Compute harmonic gradient $h_k$ with \eqref{eq:harmonic_gradient} \\
              $\theta_{k+1} = \theta_k - \eta_1 \, h_k$ \\
              {\bfseries Update critics} \\
              Compute value and cost value gradients $g_\omega$, $g_\phi$ \\
              $\omega_{k+1} = \omega_k - \eta_2 \, g_\omega$ \\
              $\phi_{k+1} = \phi_k - \eta_2 \, g_\phi$
        }
       {\bfseries return} parameters $\theta, \omega, \phi$
\end{algorithm}

\section{Simulation Verification}

In this subsection, we conduct extensive simulations in a batch of randomly generated multi-lane scenarios to demonstrate the effectiveness of our DSAC-H algorithm.

\subsection{Simulation Environment}

\begin{figure}[htbp]
  \centering
  \begin{subfigure}[b]{0.8\linewidth}
    \centering
    \includegraphics[width=\linewidth, trim={0.0cm 0.0cm 0.0cm 0.0cm}, clip]{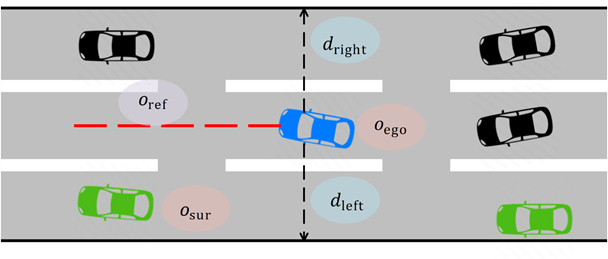}
    \caption{Observation related variables}
    \label{fig_multline_a}
  \end{subfigure}
  \vskip 0.5cm 
  \begin{subfigure}[b]{0.8\linewidth}
    \centering
    \includegraphics[width=\linewidth, trim={0.0cm 0.0cm 0.0cm 0.0cm}, clip]{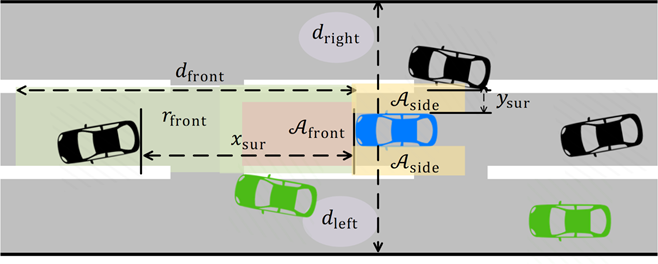}
    \caption{Reward and cost related variables}
    \label{fig_multline_b}
  \end{subfigure}
  \caption{Schematic of the multi-lane scenarios.}
  \label{fig_multline}
\end{figure}

We use a typical 6-lane bidirectional intersection with a width of each lane is 3.75 meters. The traffic flow ranges from 600 to 1200 vehicles per hour. The intersection angles are randomly generated within the range of 60 to 120 degrees. 
The control and observation variables in Figure \ref{fig_multline_a} are crucial for the simulation environment. The control variables include acceleration and angular velocity increments, denoted as $a=[\Delta a_x, \Delta \delta]$. The lane information observation values are $o_b=[d_{\text{left}}, d_{\text{right}}]$. The ego-vehicle observation values in the ego-vehicle coordinate system are $o_{\text{ego}}=[v_x, v_y, r, \delta, a_x, \delta_{-1}, a_{x, -1}]$. The reference trajectory observation values in the Frenet coordinate system (ego-ref) are $o_{\text{ref}}=[x, y, \cos(\varphi), \sin(\varphi), v] \times 4$ for the reference trajectory at 0$\rm{s}$, 0.1$\rm{s}$, 0.5$\rm{s}$, and 1$\rm{s}$. The surrounding vehicle information observation values in the ego-vehicle coordinate system (sur-ego) are $o_{\text{sur}}=[x_{\text{sur}}, y_{\text{sur}}, \cos(\varphi_{\text{sur}}), \sin(\varphi_{\text{sur}}), v_{\text{sur}}, L_{\text{sur}}, W_{\text{sur}}, \text{mask}] \times M$, where $M$ is the number of surrounding vehicles, and $\text{mask}$ indicates whether the corresponding slot contains a valid vehicle (1) or a placeholder (0) for padding.

The reward function is designed to balance trajectory tracking, control effort, comfort, and safety, based on key aspects of human-like and efficient driving. As shown in Figure~\ref{fig_multline_b}, the trajectory tracking-related reward is defined as $r_{\text{trac}}=-\rho_y \cdot h(y_{\text{ref}})-\rho_v \cdot h(v_{\text{ref}})-\rho_{\varphi} \cdot h(\varphi_{\text{ref}})$, where $h(\cdot)$ denotes the Huber loss. This term encourages the agent to follow the desired lateral position, speed, and heading, promoting accurate trajectory following. The control-related reward is $r_{\text{act}}=-\rho_\text{acc} \cdot a_x^2 - \rho_{\delta} \cdot \delta^2$, which discourages excessive control inputs to improve energy efficiency and maintain smooth steering and acceleration behavior. The comfort-related reward is $r_{\text{comf}}=-\rho_r \cdot h(r)-\rho_\text{acc} \cdot a^2 - \rho_{\delta} \cdot \delta^2$, further emphasizing passenger comfort by penalizing rapid changes in dynamics such as yaw rate and jerky control. While both $r_{\text{act}}$ and $r_{\text{comf}}$ involve control signals, they target different goals: one minimizes input effort, while the other enhances perceived ride quality. Finally, a constant survival reward of $r_{\text{live}}=12$ per time step is used to encourage longer episode durations and avoid premature terminations, which indirectly promotes safety.

Additionally, the safety-related costs include safe following distance $c_{\text{front}}=\rho_{\text{ft}} \cdot (1-\tanh(x_{\text{sur}}/(v_x \Delta t_{\text{ft}})))$ for $0 \leq x_{\text{sur}} \leq d_{\text{front}}$ and $|y_{\text{sur}}| \leq d_{\text{side}}$, 
safe space $c_{\text{space}}=\rho_{\text{fs}} \cdot (1-\tanh(x_{\text{sur}}/d_{\text{st}}))$ for $o_{\text{sur}} \in \mathcal{A}_{\text{front}}$ and $\rho_{\text{ss}} \cdot (1-\tanh(y_{\text{sur}}/d_{\text{ss}}))$ for $o_{\text{sur}} \in \mathcal{A}_{\text{side}}$, and safe boundary $c_{\text{b}}=\rho_{\text{b}} \cdot (1-\tanh(\min(d_{\text{left}}, d_{\text{right}})/d_{\text{b}}))$. 
The extra penalty for collisions and out-of-area events are 100 and 400, respectively. The mentioned reward or cost weights are listed in the Table \ref{tab_rew_para}. All values are in international units by default.


\begin{table}[h]
  \begin{center}
    \caption{Hyper-parameters of reward and cost.}
    \label{tab_rew_para}
    \begin{tabular}{l l} 
    \midrule
    \midrule
        \textbf{Symbol}  & \textbf{Value} \\ 
        \midrule
        $\Delta a_x$ & [-0.25, 0.25]\\
        $\Delta \delta$  & [-0.0065, 0.0065]\\
        $a_x$  & [-1.5, 0.8] \\
        $\delta$  & [-0.065, 0.065] \\
        \midrule
        $\rho_y$ & 2.5 \\
        $\rho_v$ & 0.4 \\
        $\rho_{\varphi}$ & 0.3 \\
        $\rho_r$ & 0.3 \\
        $\rho_{\rm{acc}}$ & 0.2 \\
        $\rho_{\delta}$ & 0.15 \\
        \midrule
        $M$ & 8 \\
        $\Delta t_{\rm{ft}}$ & 0.5 \\
        $\rho_{\rm{ft}}$ & 5.0 \\
        $\rho_{\rm{fs}}$ & 5.0 \\
        $\rho_{\rm{ss}}$ & 1.0 \\
        $\rho_{\rm{b}}$ & 1.0 \\
        $d_{\rm{front}}$ & 50.0 \\
        $d_{\rm{side}}$ & 1.8 \\
        $d_{\rm{st}}$ & 12.0 \\
        $d_{\rm{ss}}$ & 2.0 \\
        $d_{\rm{b}}$ & 1.8 \\
        $\lambda$ & 1.0 \\
    \midrule
    \midrule
    \end{tabular}
  \end{center}
\end{table}

\begin{figure}[htbp]
  \centering
  \begin{subfigure}{0.85\columnwidth}
    \includegraphics[width=\linewidth]{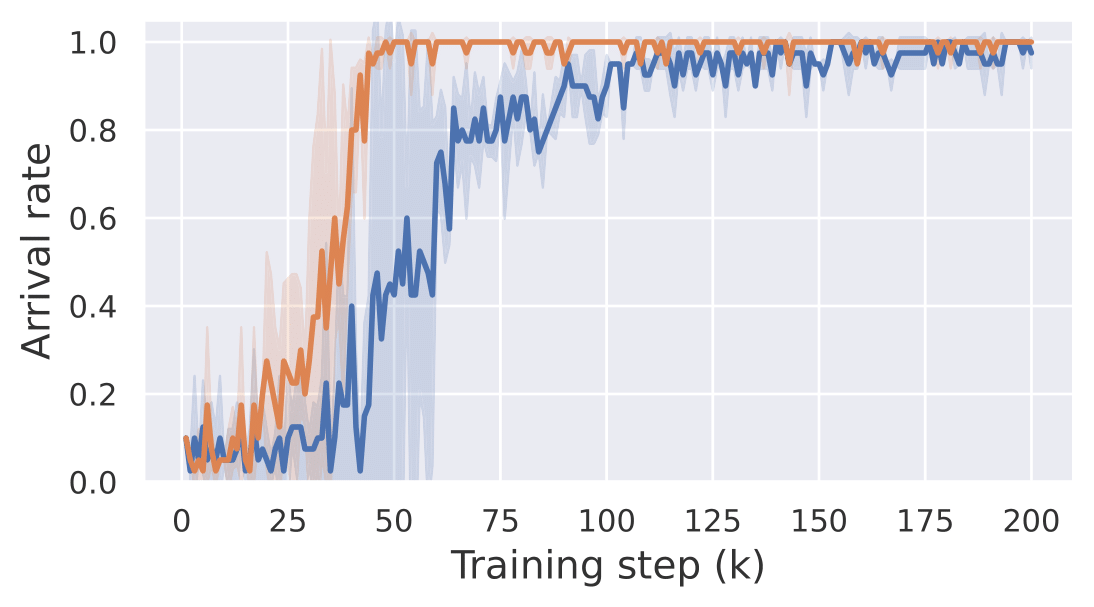} 
    \caption{Arrival rate}
    \label{fig:arr}
  \end{subfigure}
  \begin{subfigure}{0.85\columnwidth}
    \includegraphics[width=\linewidth]{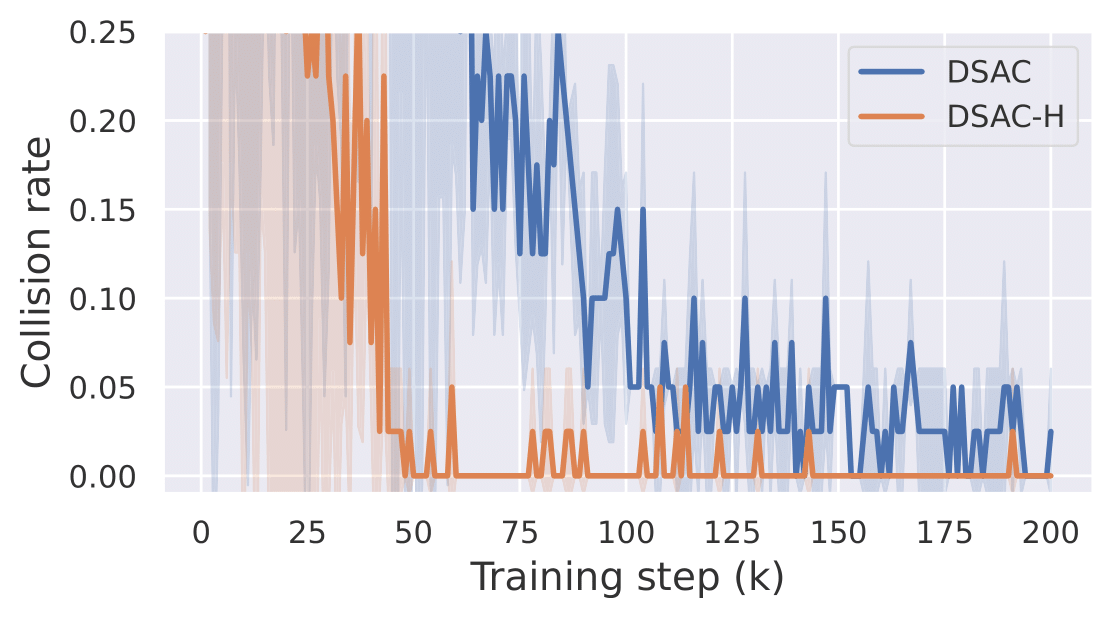}
    \caption{Collision rate}
    \label{fig:col}
  \end{subfigure}
  \caption{Training curves in the multi-lane scenarios.}
  \label{fig:curves}
\end{figure}

\begin{table}[h]
  \begin{center}
    \caption{Training hyper-parameters.}
    \label{tab_train_para}
    \begin{tabular}{l l} 
    \midrule
    \midrule
        \textbf{Symbol}  & \textbf{Value} \\ \midrule
        Actor learning rate $\eta_1$ & $3 \times 10^{-4}$\\
        Two Critic learning rate $\eta_2$  & $1 \times 10^{-4}$\\
        Sample batch size  & 20\\
        Replay batch size  & 256\\
        Optimizer & Adam\\
        Discount factor & 0.99\\
        Number of iterations & $2 \times 10^5$\\
        Learning rate of $\alpha$ & $3 \times 10^{-4}$\\
        Expected entropy & -$\dim (\mathcal{A})$\\
        Policy update frequency & 2\\
        Target update rate $\tau$ & 0.005\\
        Harmonic constant $\rho$ & 0.9\\
        Max iteration of solving $h$ & 20\\
        \midrule
        \midrule
    \end{tabular}
  \end{center}
\end{table}

\begin{figure*}[thbp]
\centering
\noindent\makebox[\textwidth][c]{
\begin{subfigure}[]{0.19\textwidth}
  \centering
  \includegraphics[width=3.6cm, trim={1cm 0.5cm 1cm 0.5cm}, clip]{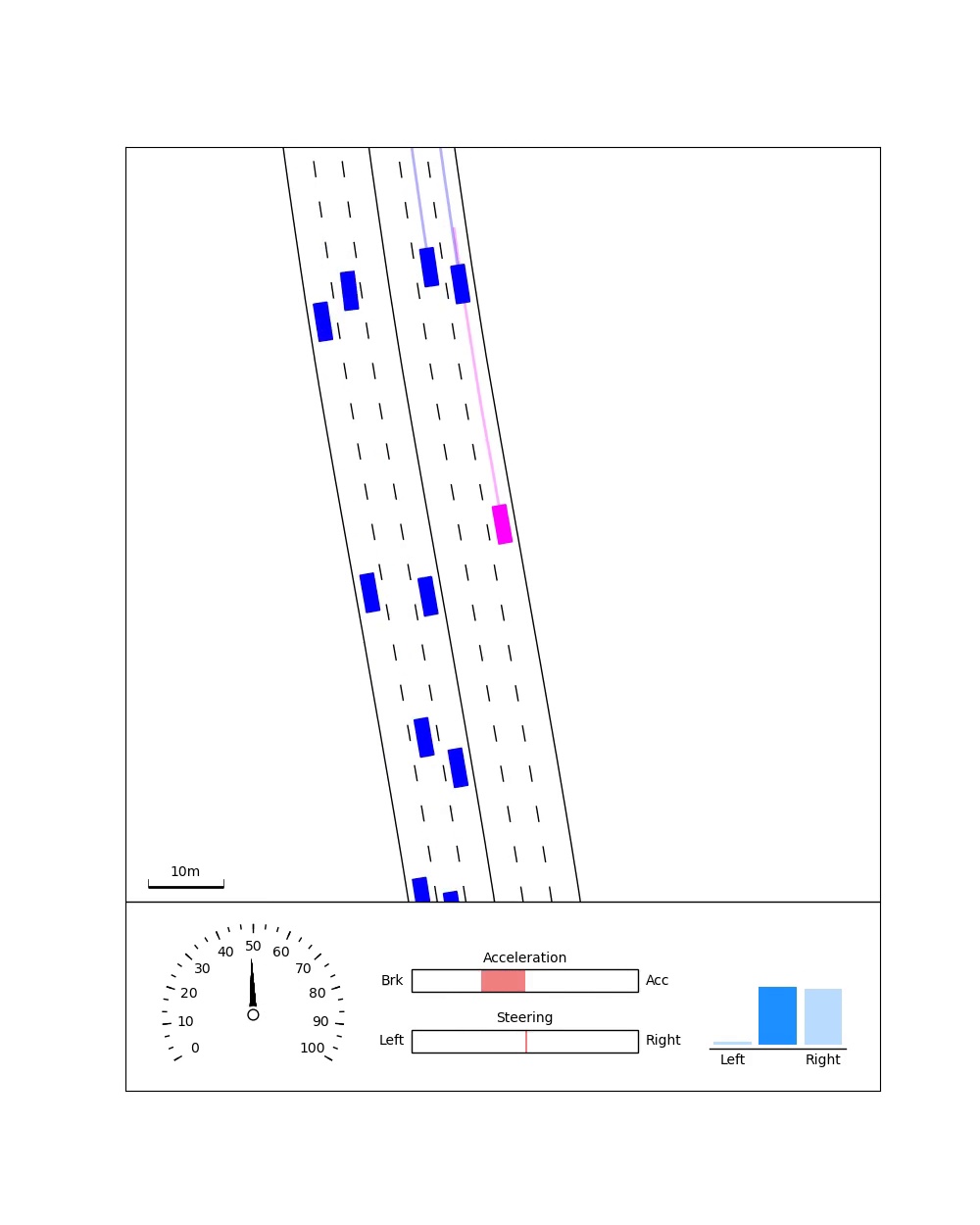}
  \caption{$t$ = 21.3$~\rm{s}$}
  \label{3a}
\end{subfigure}
\begin{subfigure}[]{0.19\textwidth}
  \centering
  \includegraphics[width=3.6cm, trim={1cm 0.5cm 1cm 0.5cm}, clip]{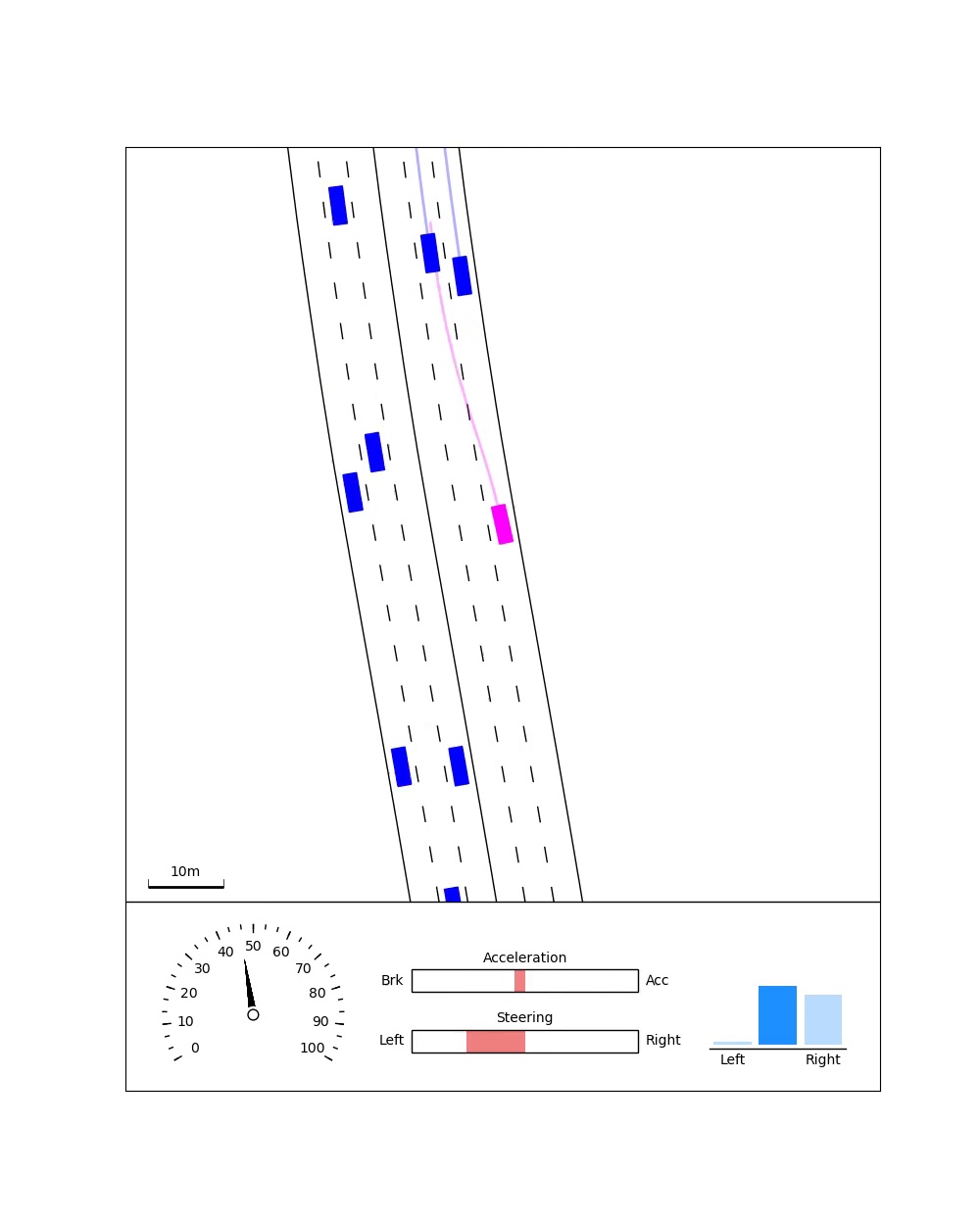}
  \caption{$t$ = 22.1$~\rm{s}$}
  \label{3b}
\end{subfigure}
\begin{subfigure}[]{0.19\textwidth}
  \centering
  \includegraphics[width=3.6cm, trim={1cm 0.5cm 1cm 0.5cm}, clip]{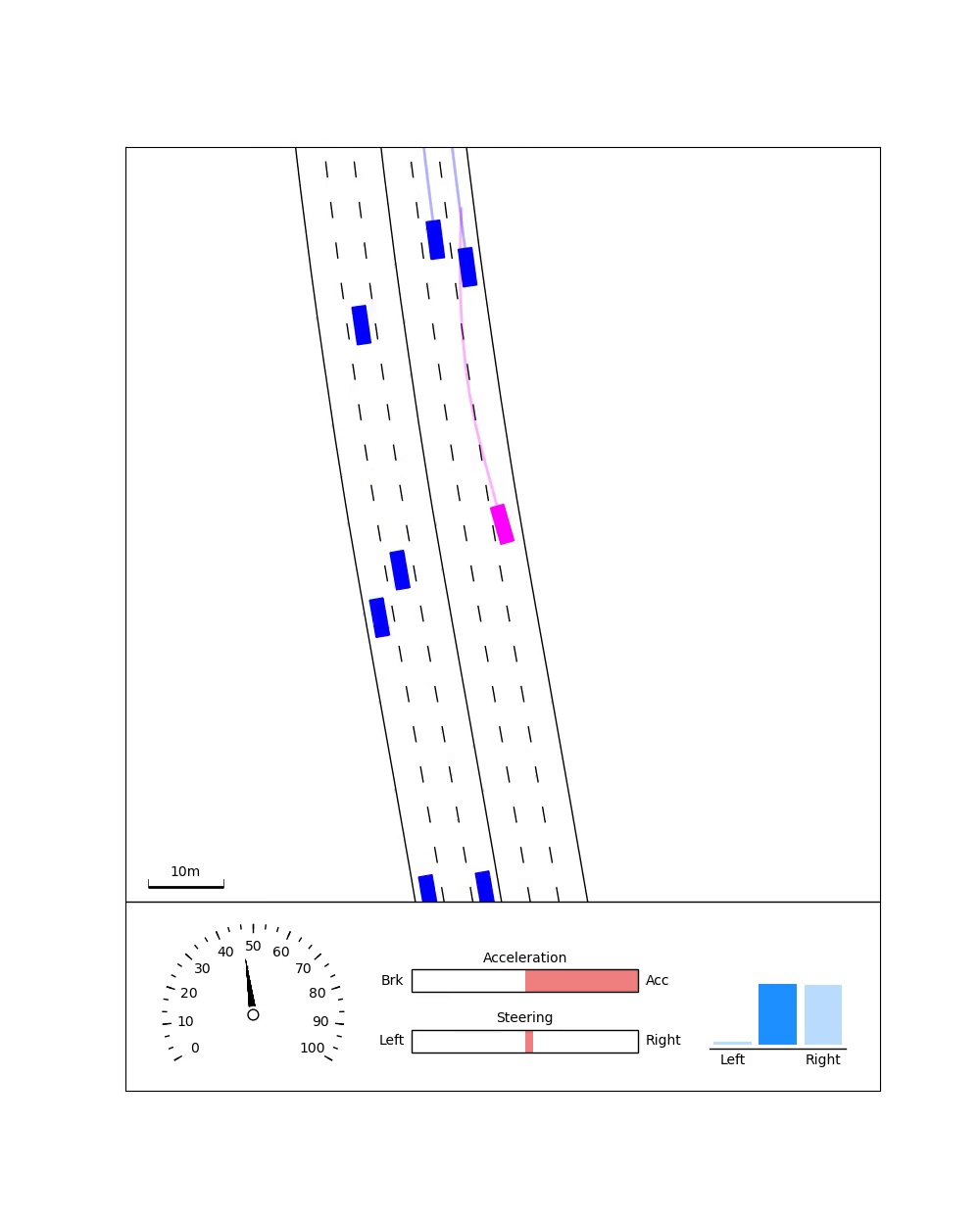}
  \caption{$t$ = 22.7$~\rm{s}$}
  \label{3c}
\end{subfigure}
\begin{subfigure}[]{0.19\textwidth}
  \centering
  \includegraphics[width=3.6cm, trim={1cm 0.5cm 1cm 0.5cm}, clip]{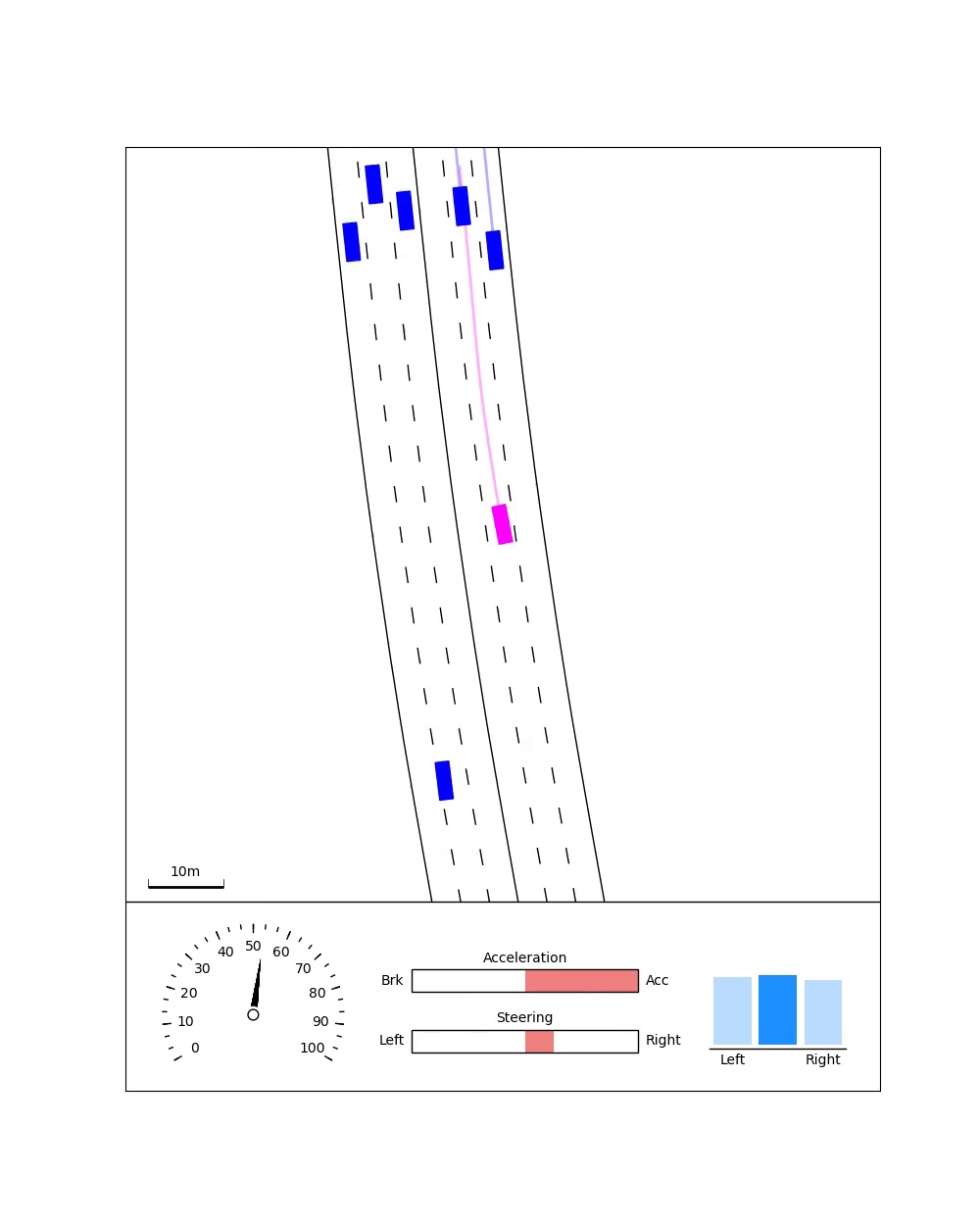}
  \caption{$t$ = 24.9$~\rm{s}$}
  \label{3d}
\end{subfigure}
\begin{subfigure}[]{0.19\textwidth}
  \centering
  \includegraphics[width=3.6cm, trim={1cm 0.5cm 1cm 0.5cm}, clip]{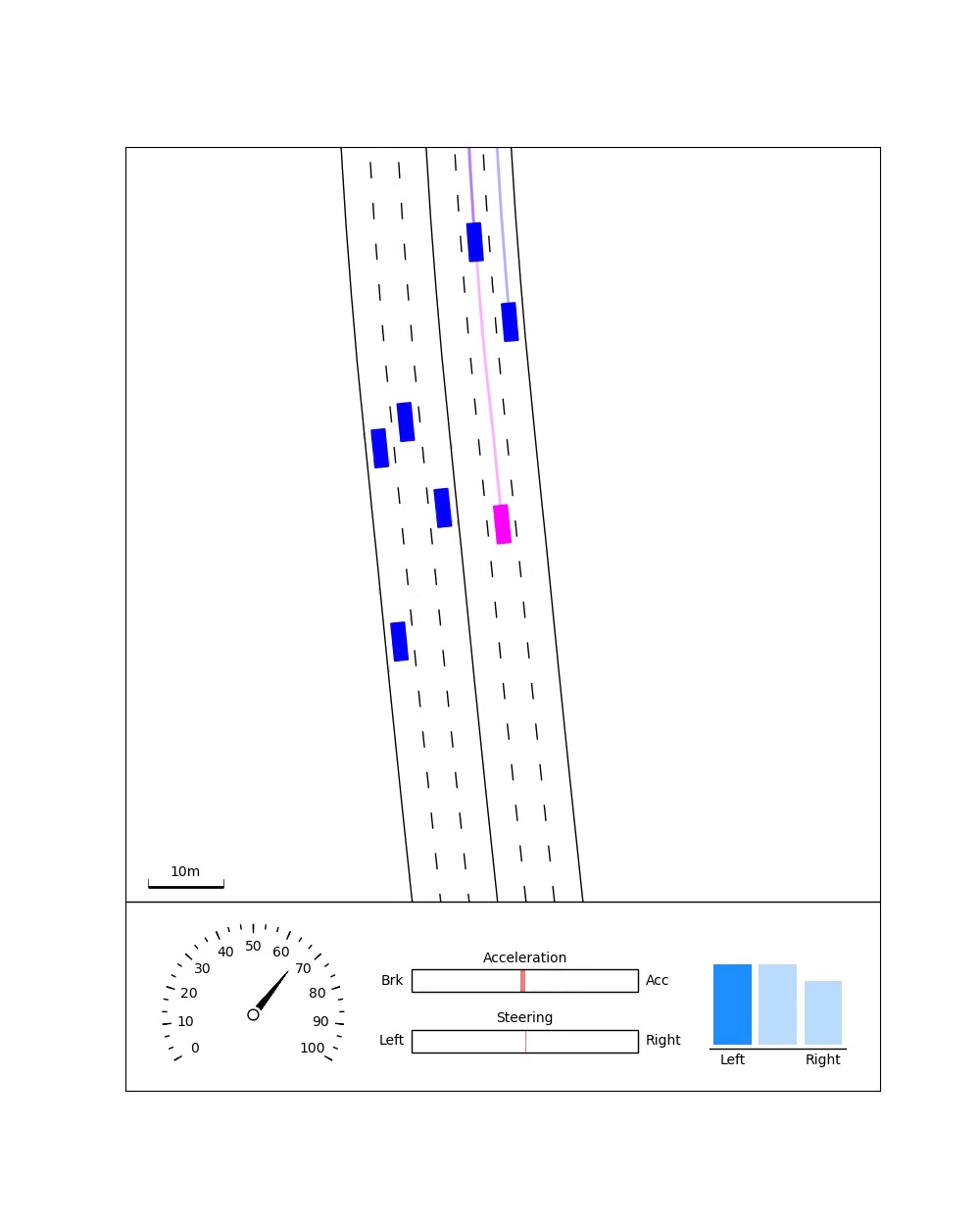}
  \caption{$t$ = 29.6$~\rm{s}$}
  \label{3e}
\end{subfigure}
}
\noindent\makebox[\textwidth][c]{
\begin{subfigure}[]{0.19\textwidth}
  \centering
  \includegraphics[width=3.6cm, trim={1cm 0.5cm 1cm 0.5cm}, clip]{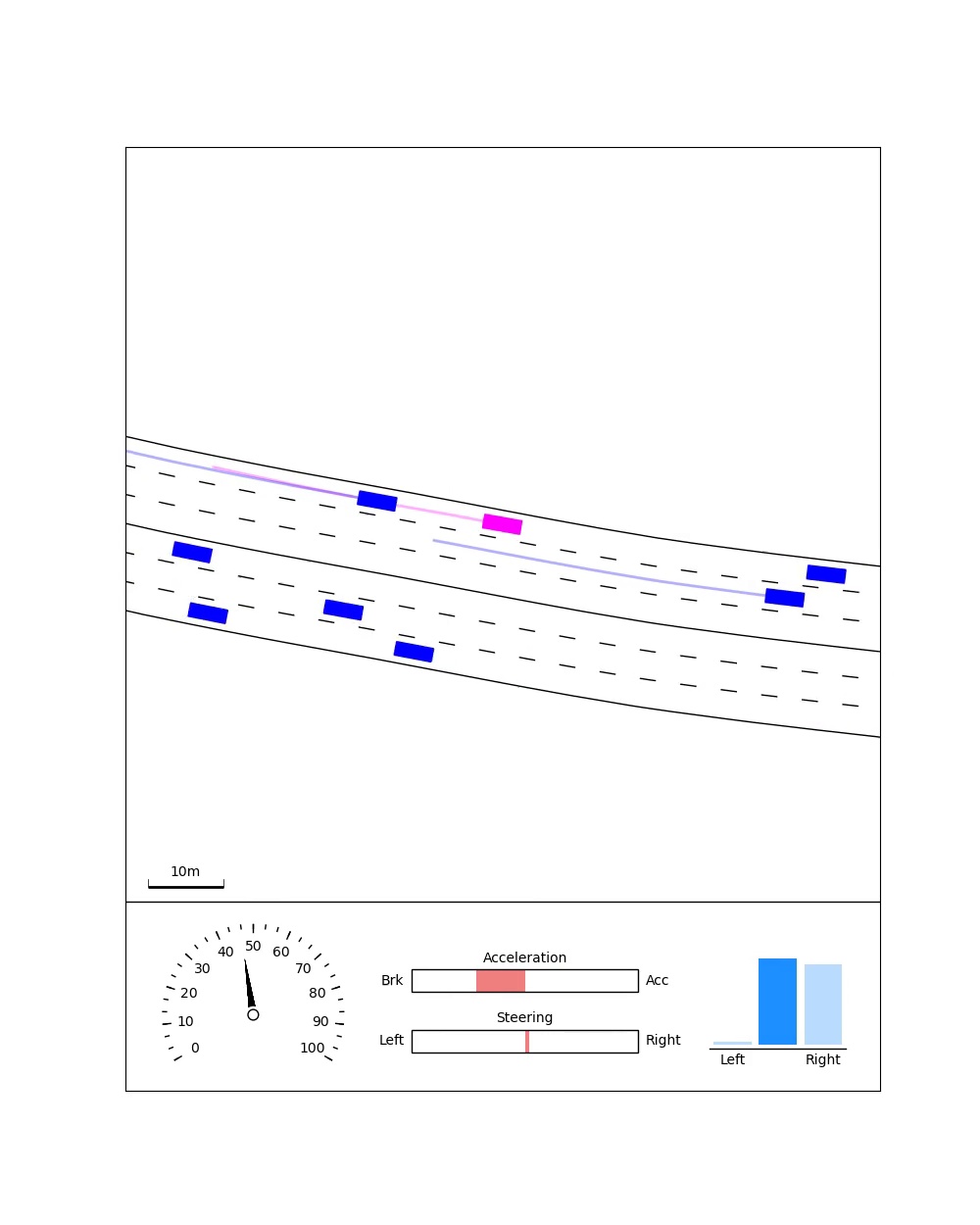}
  \caption{$t$ = 3.9$~\rm{s}$}
  \label{3f}
\end{subfigure}
\begin{subfigure}[]{0.19\textwidth}
  \centering
  \includegraphics[width=3.6cm, trim={1cm 0.5cm 1cm 0.5cm}, clip]{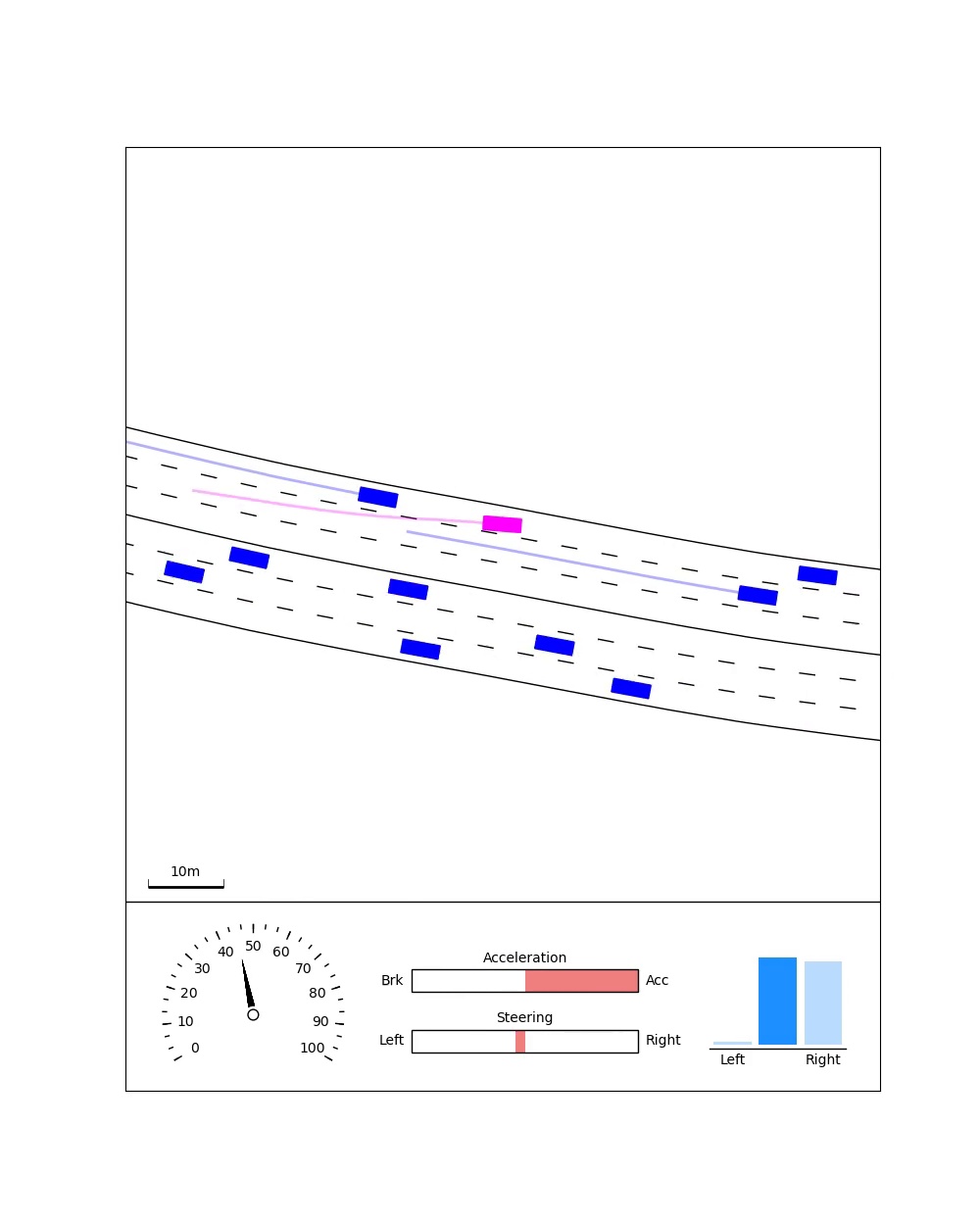}
  \caption{$t$ = 5.0$~\rm{s}$}
  \label{3g}
\end{subfigure}
\begin{subfigure}[]{0.19\textwidth}
  \centering
  \includegraphics[width=3.6cm, trim={1cm 0.5cm 1cm 0.5cm}, clip]{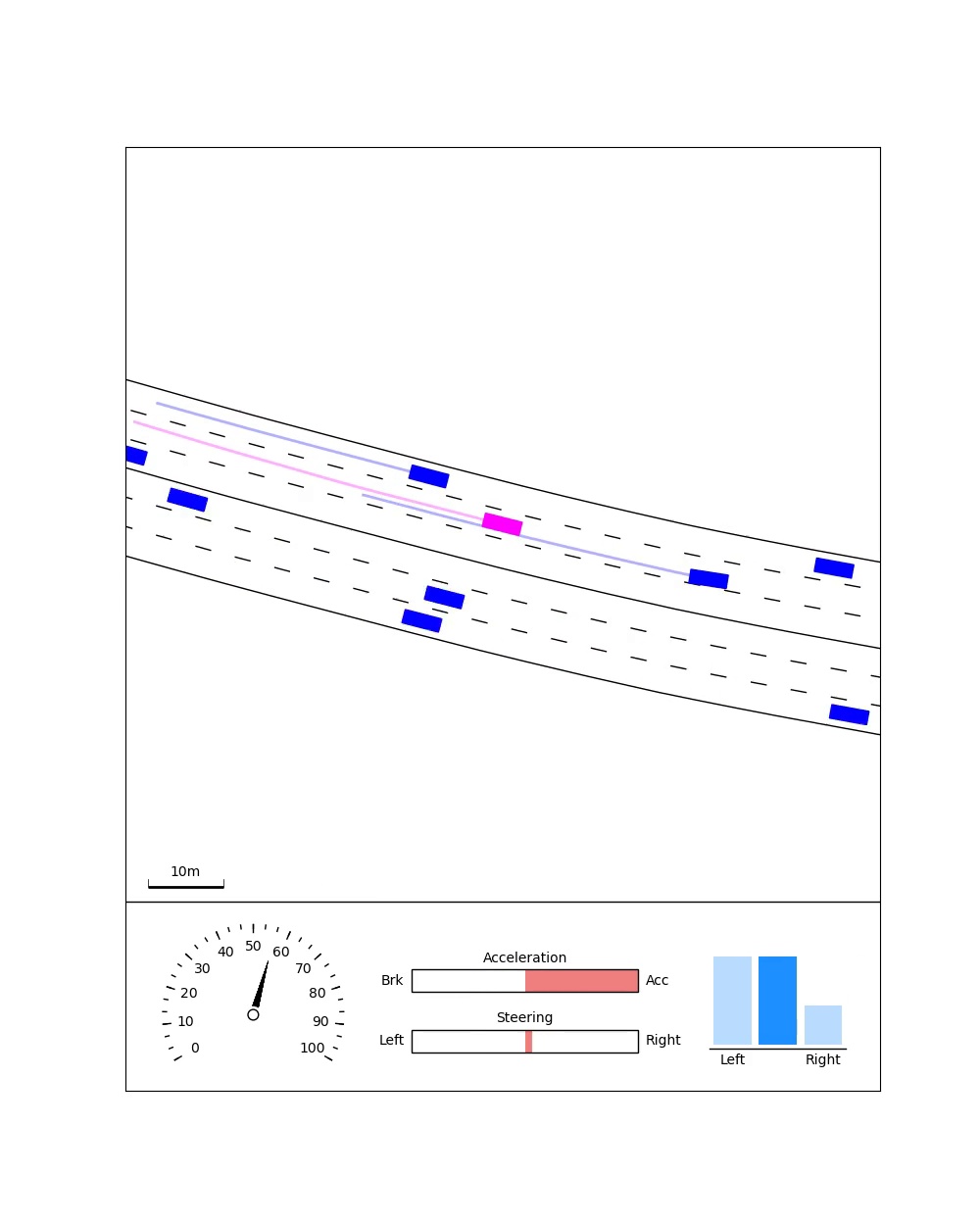}
  \caption{$t$ = 8.9$~\rm{s}$}
  \label{3h}
\end{subfigure}
\begin{subfigure}[]{0.19\textwidth}
  \centering
  \includegraphics[width=3.6cm, trim={1cm 0.5cm 1cm 0.5cm}, clip]{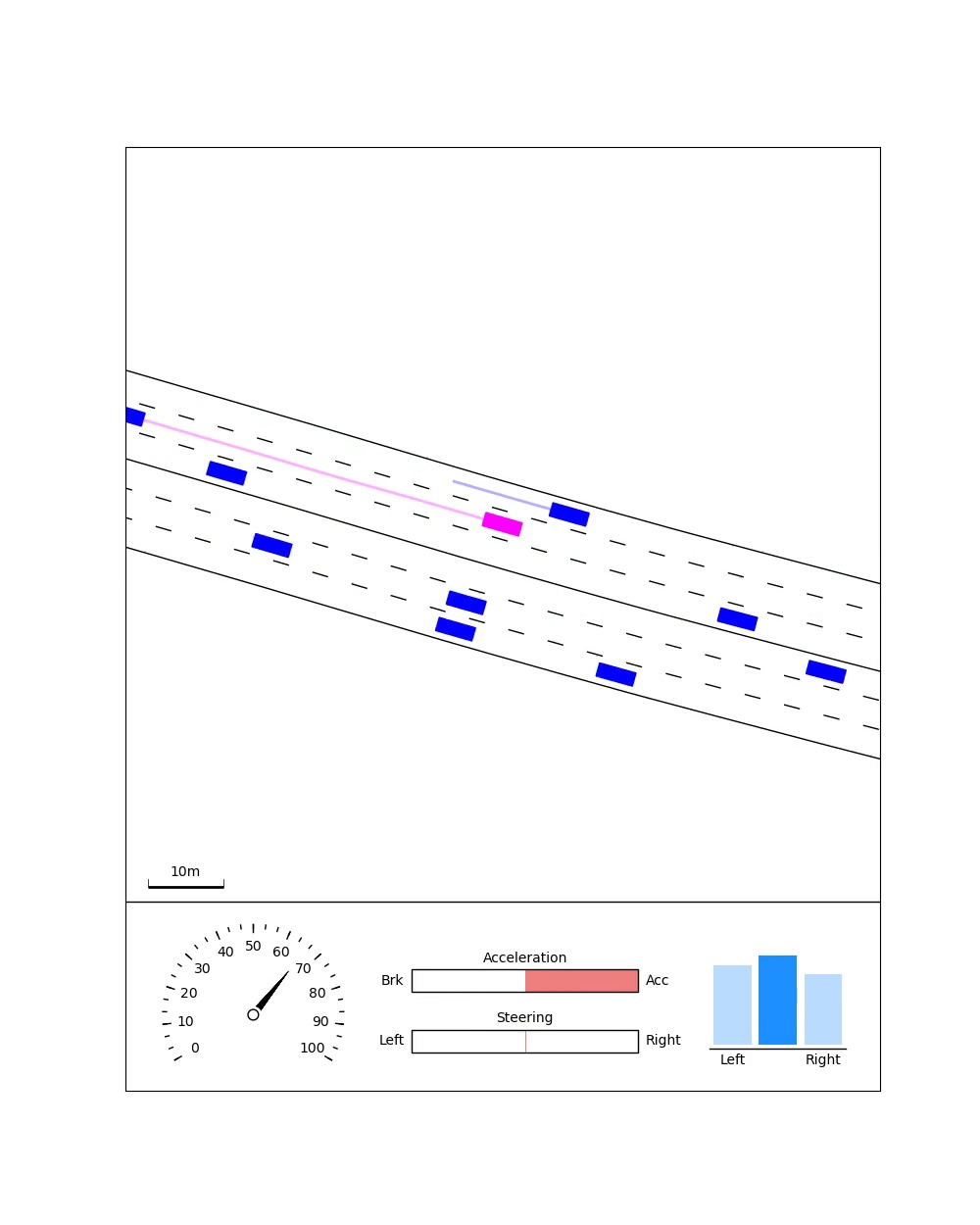}
  \caption{$t$ = 12.3$~\rm{s}$}
  \label{3i}
\end{subfigure}
\begin{subfigure}[]{0.19\textwidth}
  \centering
  \includegraphics[width=3.6cm, trim={1cm 0.5cm 1cm 0.5cm}, clip]{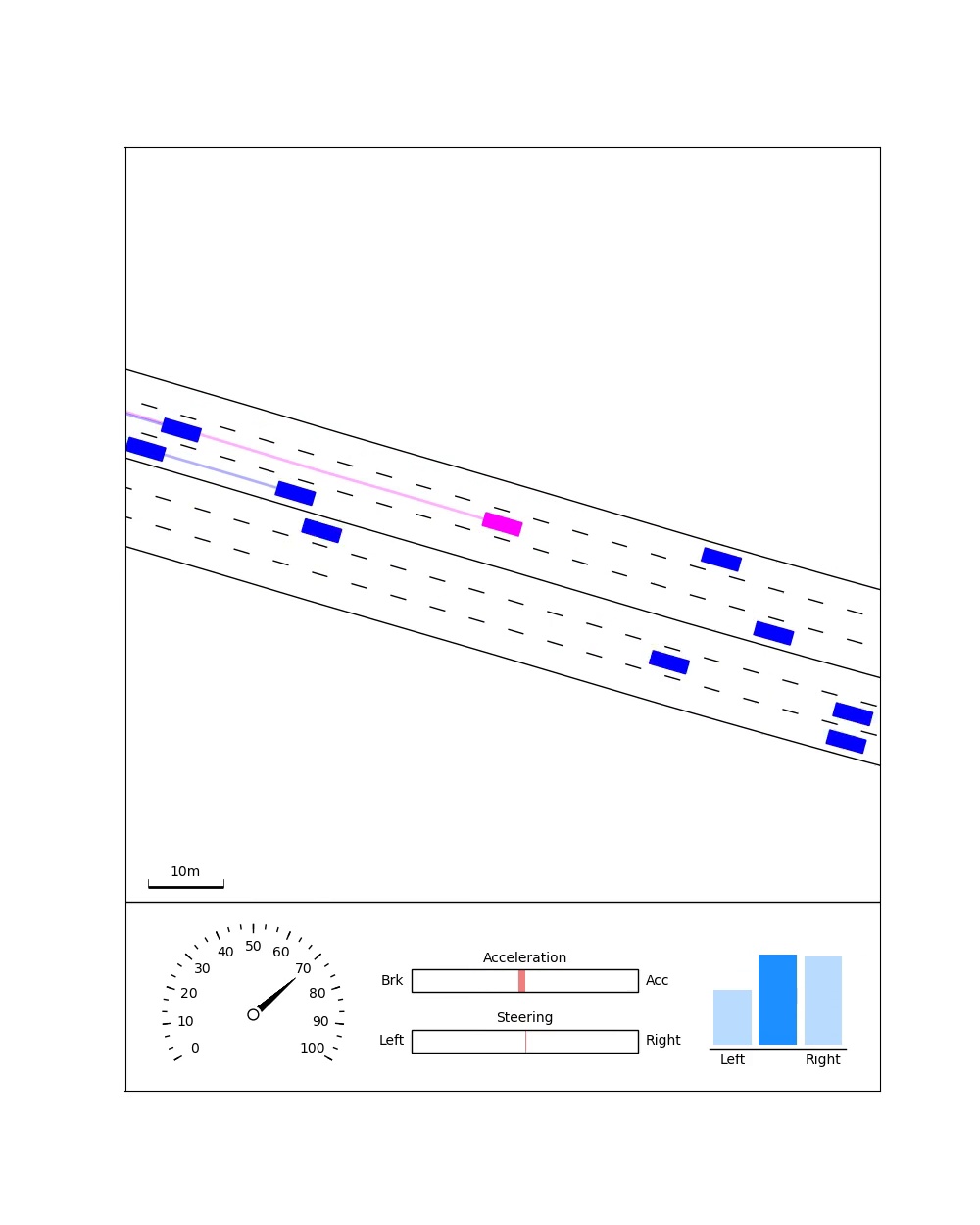}
  \caption{$t$ = 13.9$~\rm{s}$}
  \label{3j}
\end{subfigure}
}
\caption{Visualization of two typical cases in multi-lane scenarios. Case 1 (the first row) involves following and lane-changing behaviors. Case 2 (the second row) involves an overtaking behavior.}
\label{fig:visualization}
\end{figure*}

\subsection{Training Settings}

We use the GOPS software to conduct RL training \cite{wang2023gops}, and the main hyper-parameters are listed in Table \ref{tab_train_para}.
The sampling process ensures a balanced distribution of scenarios. 
The ego-vehicle's initial state is randomly selected from the traffic flow with a small Gaussian bias.
The reference trajectory is randomly chosen from the current or adjacent lanes, with random desired speed for tracking. Surrounding vehicles have random lateral position and size biases.  During replay, a hierarchical prioritized experience replay sampling method is used, with event types including collision, braking, out-of-area, and normal driving. 
The importance indicator is $pri=(Q_{r,\text{mean}}-Q_{r, \text{target}})^2$, and the sampling weight is proportional to the indicator value.
The probability of each event type is calculated based on the importance indicator and sampling weight.

\subsection{Training Curves and Visualization Demos}

Figure \ref{fig:curves} compares the arrival rate and collision rate in the training process between DSAC and DSAC-H algorithms. As illustrated in Figure \ref{fig:arr}, the arrival rate of DSAC-H increases more rapidly and stabilizes at a higher level compared to DSAC, indicating faster convergence and better performance in terms of successful task completion. Meanwhile, Figure \ref{fig:col} shows that the collision rate of DSAC-H remains consistently lower than that of DSAC throughout the training steps, demonstrating its enhanced safety and stability. These trends clearly highlight the superiority of DSAC-H in achieving a balance between efficiency and safety.


Two typical simulation cases of the DSAC-H algorithm in the multi-lane environment are visualized in Figure 3. Case 1 is the following and lane-changing tasks shown in the first 5 subfigures. Figure 3a shows the vehicle starting, Figure 3b shows the vehicle beginning to turn, Figures 3c and 3d show the vehicle accelerating to complete the lane change, and Figure 3e shows the vehicle finally reaching the reference speed and maintaining a stable following distance. Case 2 is the overtaking tasks shown in the last 5 subfigures. Figure 3f shows the vehicle starting, Figures 3g, 3h and 3i show the vehicle turning and accelerating to overtake the slower vehicle, and Figure 3j shows the vehicle finally reaching the reference speed and maintaining a safe following distance. The three vertical bars in the bottom right corner of each subfigure represent the agent’s action preferences at each time step. They correspond to the values for taking the actions: lane change to the left, lane keeping, and lane change to the right, respectively. DSAC-H maintains a safe following distance, accurately tracks the reference trajectory, executes smooth lane changes, and safely overtakes vehicles while adhering to safety constraints.

\subsection{Statistical Results Analysis}

A total of 50 scenarios not included in the training set were selected for simulation tests, covering a total driving distance of 600 kilometers. Throughout the testing process, no collisions involving the ego vehicle occurred, indicating that the driving control strategy demonstrates a high level of safety. To further evaluate the driving performance of the control strategy, this section analyzes the state errors, states, and control variables of the ego vehicle during driving, and presents the distribution of these variables.

\begin{figure}[htbp]
  \centering
  \begin{subfigure}{0.48\columnwidth}
    \includegraphics[width=\linewidth]{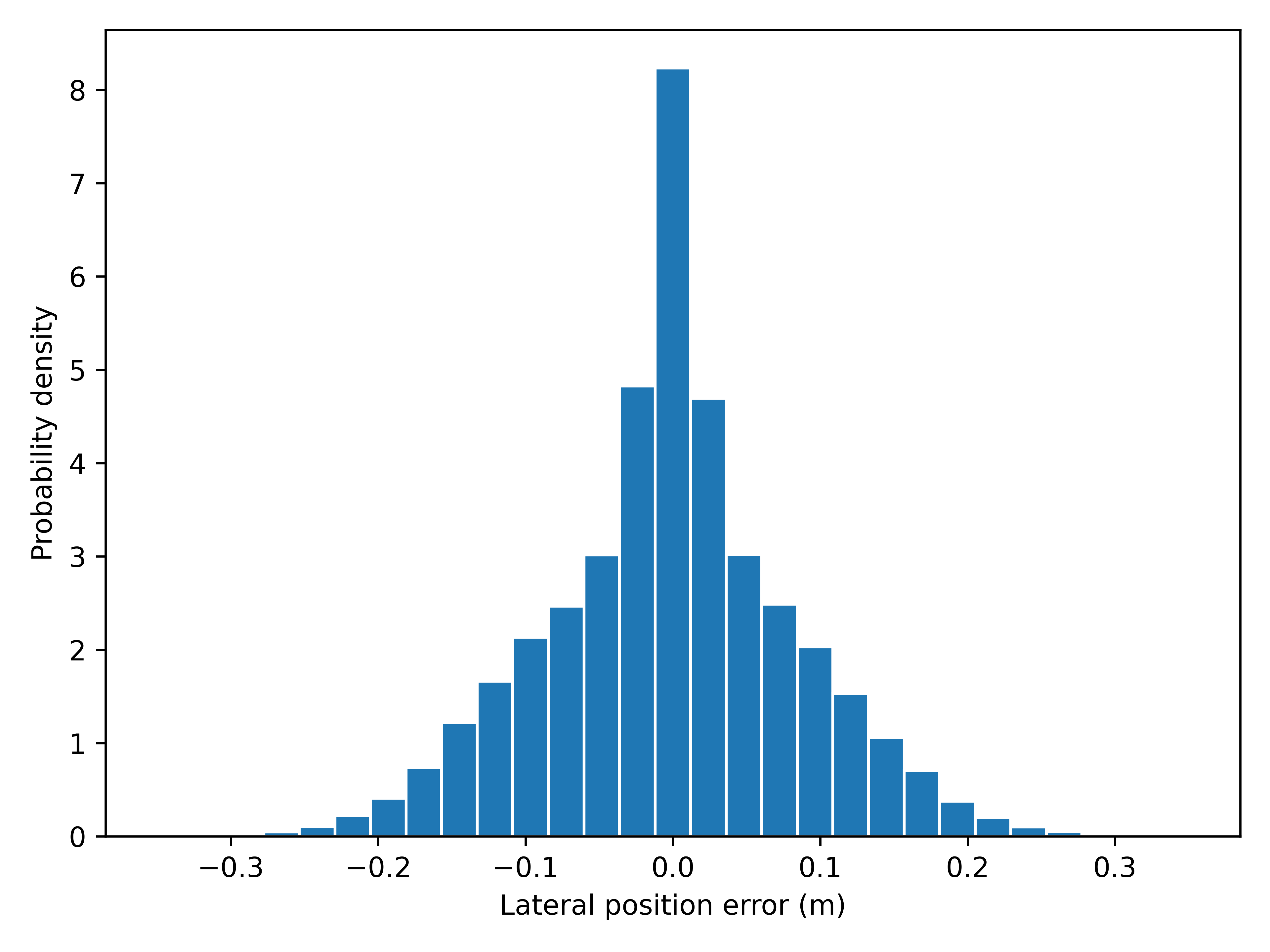} 
    \caption{Lateral position}
    \label{fig:ey}
  \end{subfigure}
  \hfill  
  \begin{subfigure}{0.48\columnwidth}
    \includegraphics[width=\linewidth]{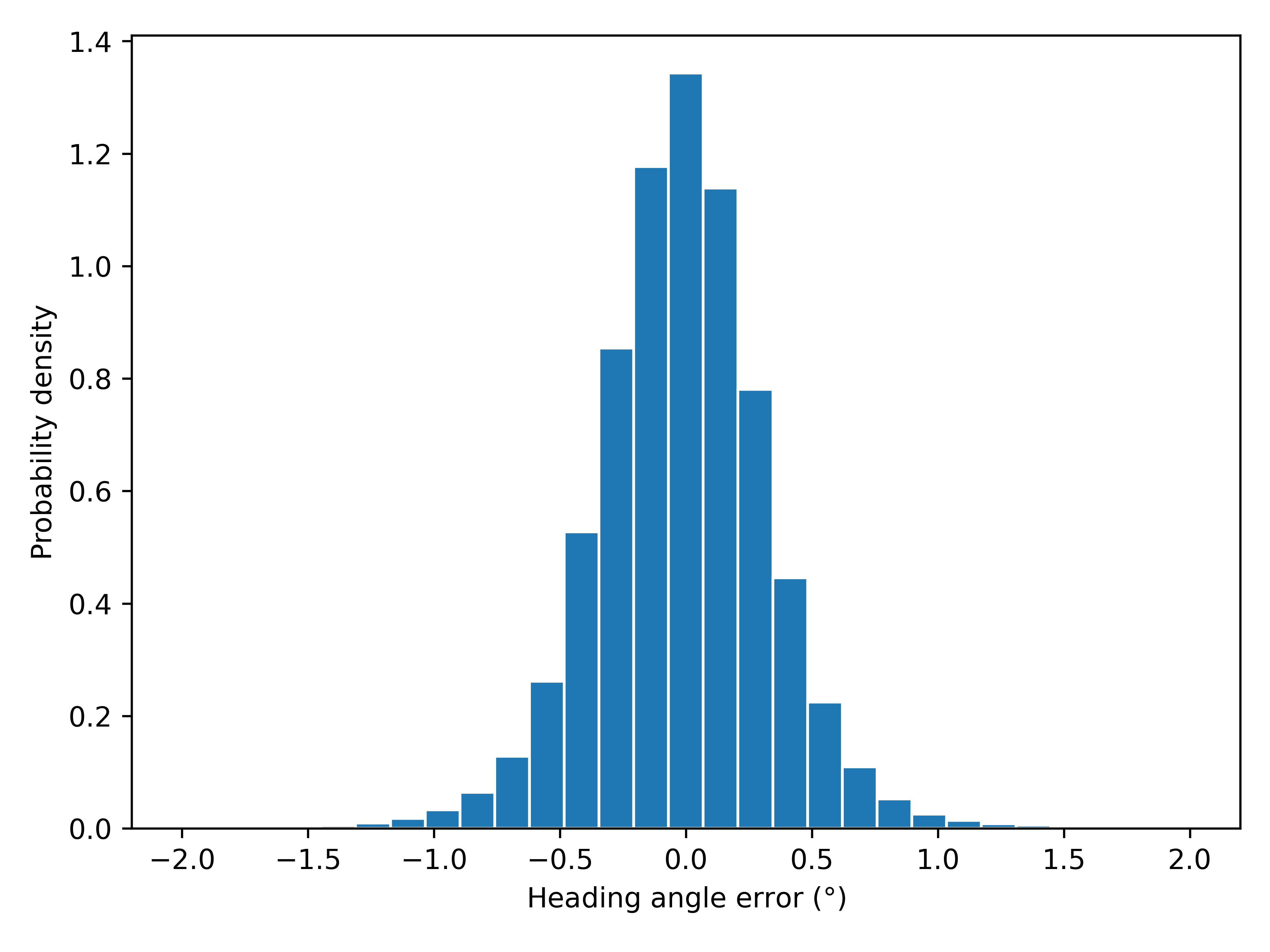} 
    \caption{Heading angle}
    \label{fig:ephi}
  \end{subfigure}
  \caption{Histogram of position tracking errors.}
\end{figure}

The distribution of the lateral position tracking error of the ego vehicle is shown in Fig.~\ref{fig:ey}, where the horizontal axis represents the lateral position tracking error, and the vertical axis denotes the probability density of the error values. It can be observed that significant probability density peaks occur around $\pm0.125\rm{m}$, and the probability density decreases sharply as the tracking error increases. This result indicates that the peaks correspond to lane-keeping behavior, where the tracking error is minimal, suggesting that the driving strategy exhibits excellent tracking performance. The tracking errors on both sides correspond to lane-changing behavior, where the errors are slightly larger but still within a small range, demonstrating that the driving strategy effectively handles lane-changing maneuvers.
The distribution of the heading angle tracking error is shown in Fig.~\ref{fig:ephi}. The error values follow a normal distribution and fall within a range of $\pm{1.5}^{\circ}$, indicating that the heading angle errors are minimal and that the driving strategy provides good tracking performance.

\begin{figure}[htbp]
  \centering
  \begin{subfigure}{0.48\columnwidth}
    \includegraphics[width=\linewidth]{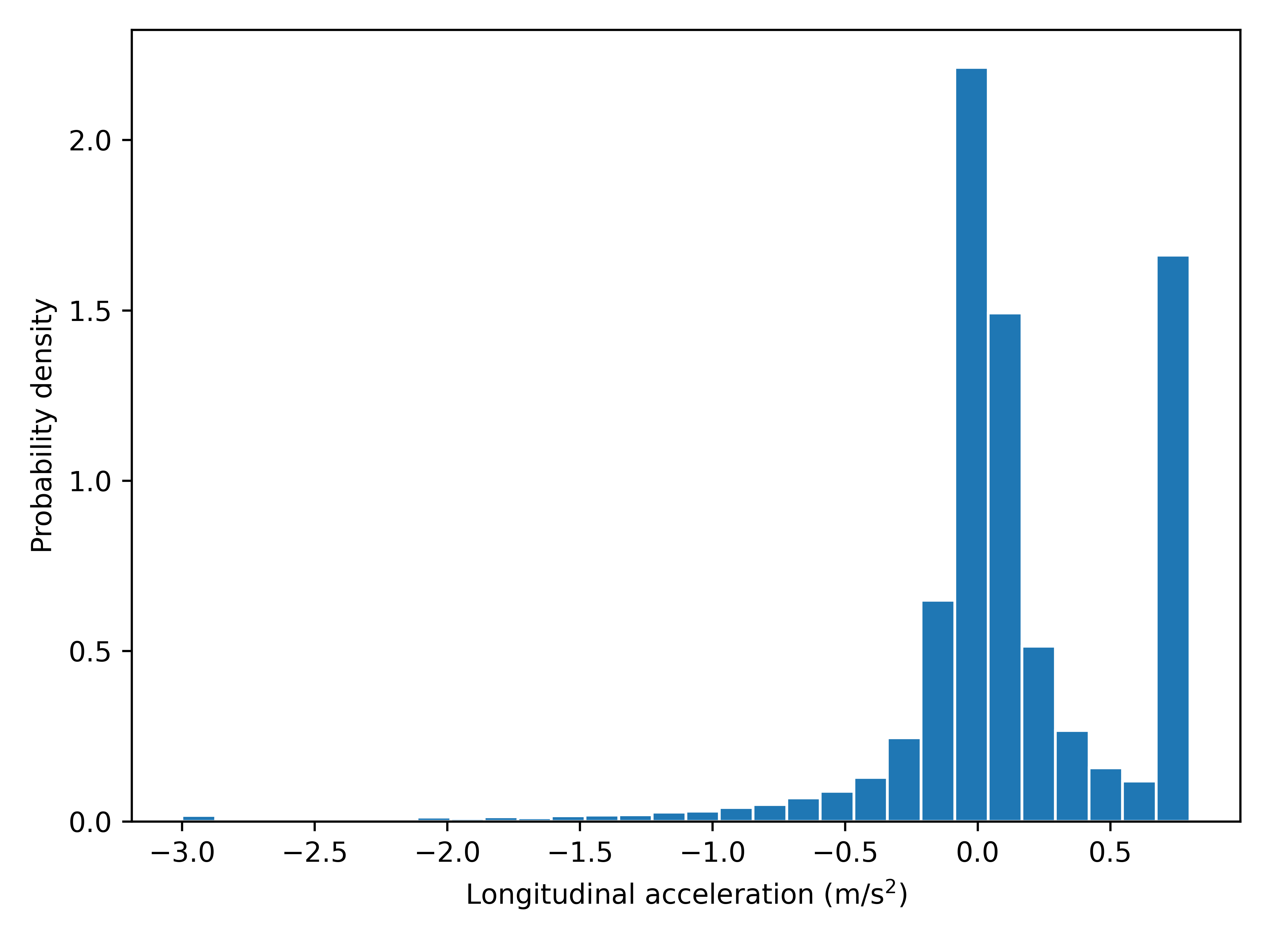} 
    \caption{longitudinal acceleration}
    \label{fig:acc}
  \end{subfigure}
  \hfill  
  \begin{subfigure}{0.48\columnwidth}
    \includegraphics[width=\linewidth]{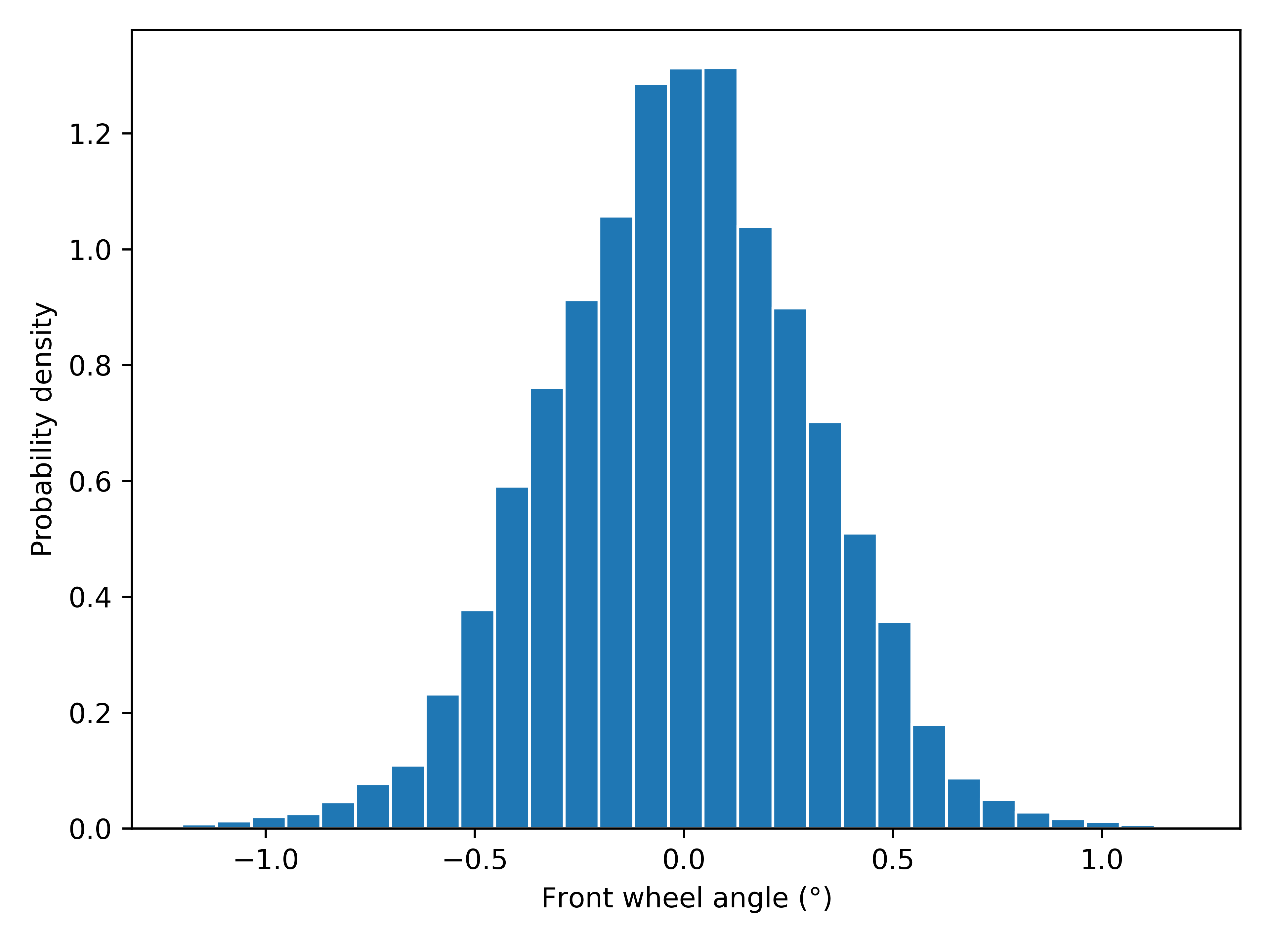} 
    \caption{Lateral steering angle}
    \label{fig:steer}
  \end{subfigure}
  \caption{Histogram of vehicle control actions.}
\end{figure}

The longitudinal acceleration distribution of the ego vehicle is presented in Fig.~\ref{fig:acc}. Two peaks are observed: one near $0.8~\rm{m/s^2}$, corresponding to the ego vehicle accelerating when its longitudinal speed is below the desired speed, and another near 0, representing steady driving at the desired speed. Since the ego vehicle's initial longitudinal speed in the test scenarios is distributed between zero and the desired speed, the statistics show a higher frequency of acceleration events.
The distribution of the front wheel steering angle is shown in Fig.~\ref{fig:steer}. The steering angle values follow a normal distribution and fall within a range of $\pm{1.2}^{\circ}$, indicating that the front wheel angles are small. This result demonstrates that the driving strategy enables the ego vehicle to smoothly follow the reference trajectory, providing good stability.

\begin{figure}[htbp]
  \centering
  \begin{subfigure}{0.48\columnwidth}
    \includegraphics[width=\linewidth]{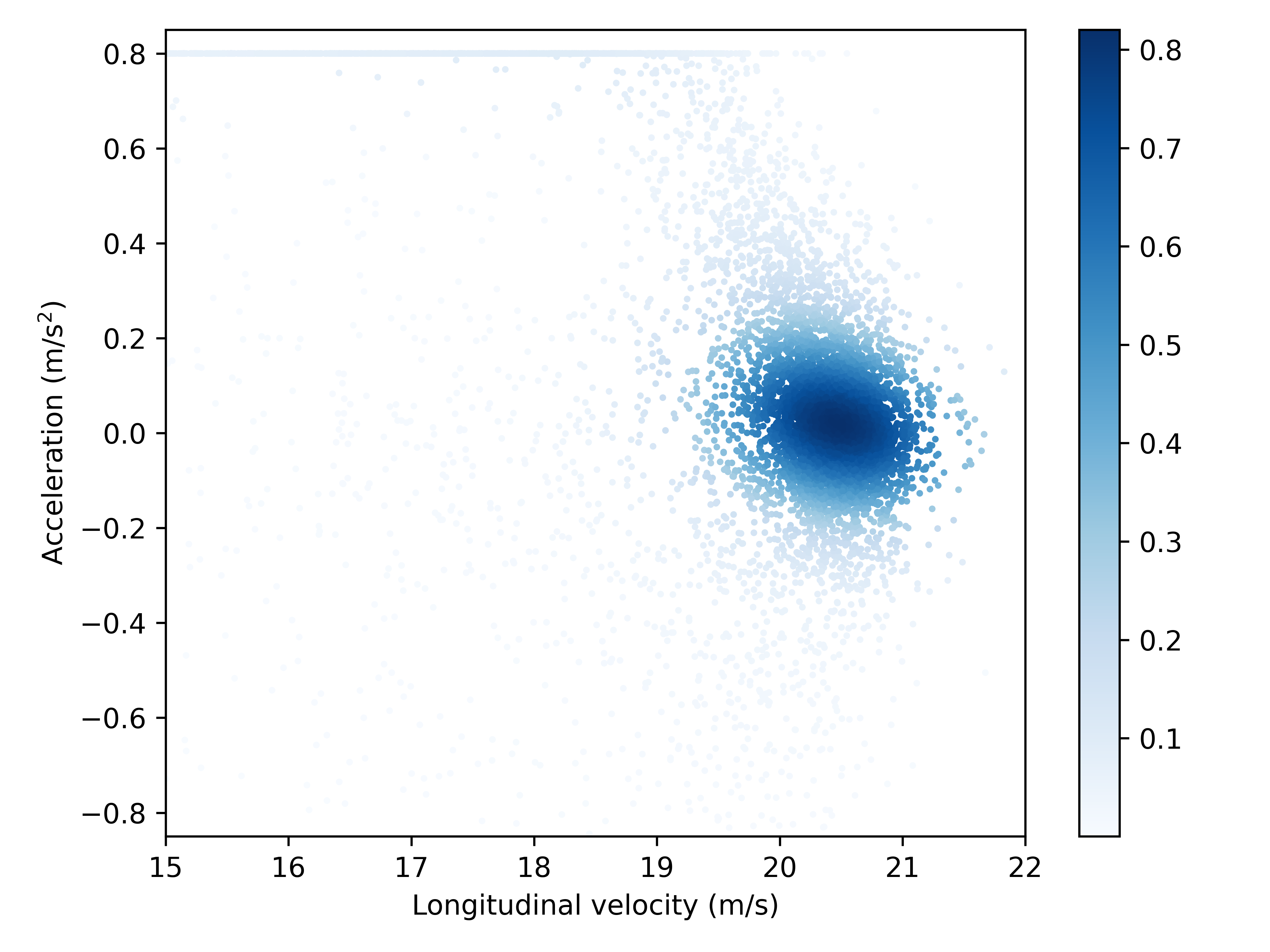} 
    \caption{Longitudinal}
    \label{fig:acc_vx}
  \end{subfigure}
  \hfill  
  \begin{subfigure}{0.48\columnwidth}
    \includegraphics[width=\linewidth]{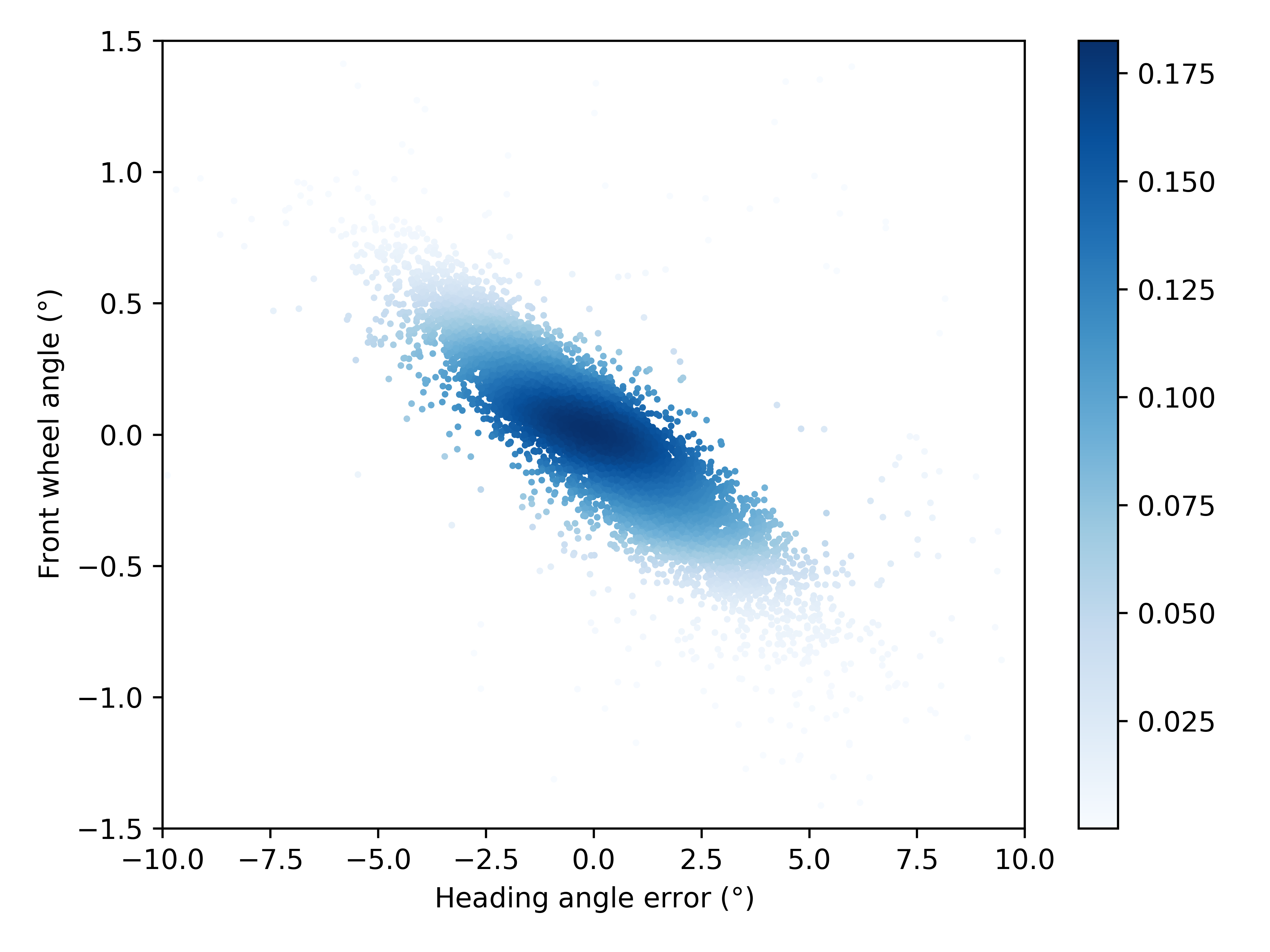} 
    \caption{Lateral}
    \label{fig:steer_ephi}
  \end{subfigure}
  \caption{Joint distributions of key states and actions.}
\end{figure}

The joint distribution of longitudinal acceleration and longitudinal speed is shown in Fig.~\ref{fig:acc_vx}. A significant probability density peak occurs near zero acceleration and the desired speed, indicating that the driving strategy effectively maintains the vehicle speed around the desired value with good stability. When the speed is below the desired value, the acceleration is positive, demonstrating that the driving strategy can promptly accelerate, achieving high driving efficiency.
The joint distribution of the front wheel steering angle and heading angle error is shown in Fig.~\ref{fig:steer_ephi}. A clear linear relationship is observed: when the heading angle error is positive, the front wheel steering angle is negative, and when the heading angle error is negative, the front wheel steering angle is positive. This result indicates that the driving strategy can adjust the heading angle of the ego vehicle in a timely manner, ensuring it follows the reference path with good tracking performance.

\section{Conclusion}  

In this paper, we propose Harmonic Policy Iteration (HPI), a safety-oriented training technique designed to address the challenges of constraint handling in RL for autonomous driving. By computing and harmonizing policy gradients associated with efficient driving and safety constraints, HPI enables more stable and balanced policy updates. Integrating HPI into the state-of-the-art DSAC algorithm, we develop DSAC-H, a safe RL algorithm capable of improving driving efficiency while maintaining near-zero safety constraint violations. Extensive simulations in multi-lane scenarios validate the effectiveness of DSAC-H, demonstrating its potential for real-world autonomous driving applications. In future work, we plan to evaluate the robustness of DSAC-H in more challenging scenarios, such as interactions with aggressive drivers or near-collision situations, to further assess its safety and generalization capabilities.


\addtolength{\textheight}{-12cm}   



\bibliographystyle{./bibliography/IEEEtran}
\bibliography{bibliography/main.bbl}

\end{document}